\documentclass[sigconf]{acmart}
\usepackage{cleveref}

\AtBeginDocument{%
  }

\copyrightyear{2026}
\acmYear{2026}
\setcopyright{cc}
\setcctype{by}
\acmConference[CAIS '26]{ACM Conference on AI and Agentic Systems}{May 26--29, 2026}{San Jose, CA, USA}
\acmBooktitle{ACM Conference on AI and Agentic Systems (CAIS '26), May 26--29, 2026, San Jose, CA, USA}
\acmDOI{10.1145/3786335.3813198}
\acmISBN{979-8-4007-2415-2/2026/05}

\begin{document}

\title[Agent-Aided Design Demo]{Agent-Aided Design for Dynamic CAD Models}

\author{Mitch Adler}
\affiliation{%
  \institution{Independent Researcher}
  \city{San Francisco}
  \state{CA}
  \country{USA}
}
\email{mnadler90@gmail.com}

\author{Matthew Russo}
\orcid{0009-0005-9685-3976}
\affiliation{%
  \institution{MIT}
  \city{Cambridge}
  \state{MA}
  \country{USA}
}
\email{mdrusso@csail.mit.edu}

\author{Michael Cafarella}
\affiliation{%
  \institution{MIT}
  \city{Cambridge}
  \state{MA}
  \country{USA}
}
\email{michjc@csail.mit.edu}

\renewcommand{\shortauthors}{Mitch Adler, Matthew Russo, and Michael Cafarella}
\newcommand{\system}{{\sc AADvark}}

%

\begin{abstract}
In the past year, researchers have created agentic systems that can design real-world CAD-style objects in a training-free setting, a new variety of system that we call \textbf{Agent-Aided Design}. These systems place an agent in a feedback loop in which it generates an assembly of CAD model(s), visualizes the assembly, and then iteratively refines its assembly based on visual and other feedback. Despite rapid progress, a key problem remains: none of these systems can build \textit{complex 3D assemblies with moving parts}. For example, no existing system can build a piston, a pendulum, or even a pair of scissors. In order for Agent-Aided Design to make a real impact in industrial manufacturing, we need a system that is capable of generating such 3D assemblies. In this paper we present a prototype of \system{}, an agentic system designed for this task. Unlike previous state-of-the-art systems, \system{} captures the dynamic part interactions with one or more degrees-of-freedom. This design decision allows \system{} to reason directly about assemblies with moving parts and can thereby achieve cross-cutting goals, including but not limited to mechanical movements. Unfortunately, current LLMs are imperfect spatial reasoners, a problem that \system{} addresses by incorporating external constraint solver tools with a specialized visual feedback mechanism. We demonstrate that, by modifying the agent's tools (FreeCAD and the assembly solver), we are able to create a strong verification signal which enables our system to build 3D assemblies with movable parts.

\end{abstract}




\maketitle


\begin{figure*}
\includegraphics[width=\textwidth]{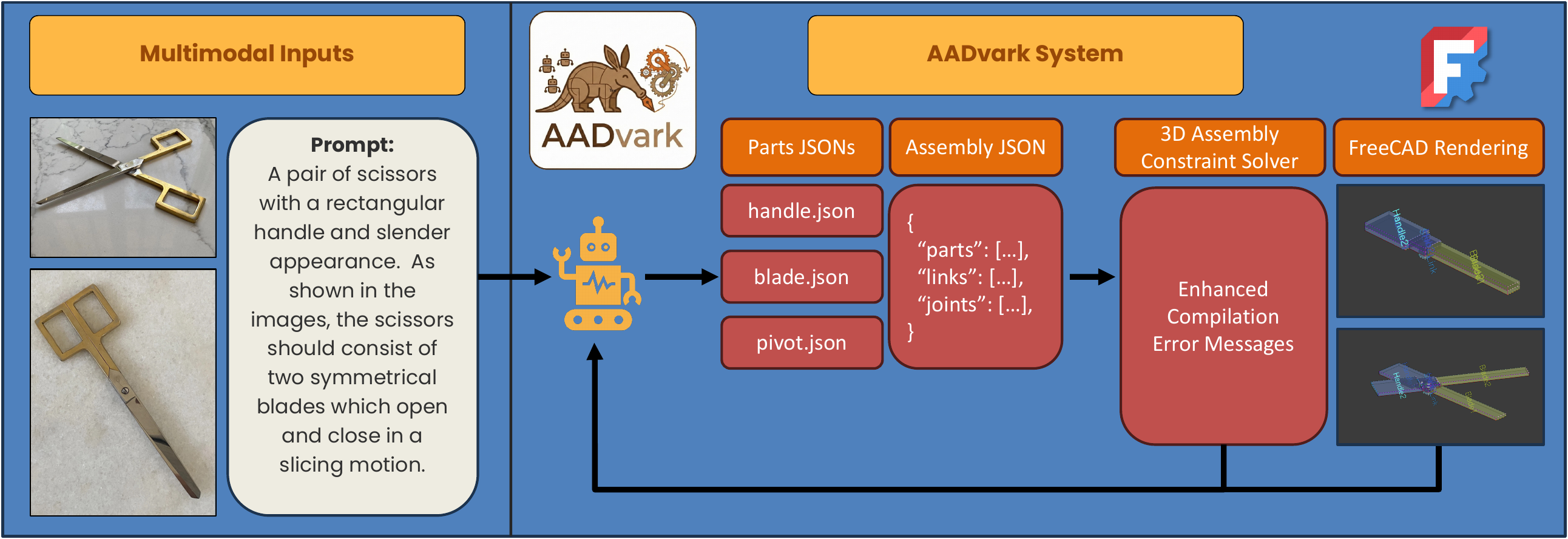}
\caption{An illustration of \system{} generating a pair of scissors. \system{} accepts one or more images and/or textual description as input. \system{} creates  JSON definitions for the parts and joints in the 3D assembly and then compiles them using a 3D assembly constraint solver. Compilation errors and intermediate renderings (generated in FreeCAD) are fed back into the agent. We modify FreeCAD and the constraint solver to provide spatial cues and informative error messages, respectively, to the agent. The agent iteratively updates the assembly until it decides the rendered object is sufficiently similar to the input(s).}

\Description{An illustration of AADvark generating a pair of scissors. AADvark accepts one or more images and/or textual description as input. AADvark creates  JSON definitions for the parts and joints in the 3D assembly and then compiles them using a 3D assembly constraint solver. Compilation errors and intermediate renderings (generated in FreeCAD) are fed back into the agent. We modify FreeCAD and the constraint solver to provide spatial cues and informative error messages, respectively, to the agent. The agent iteratively updates the assembly until it decides that the rendered object is sufficiently similar to the input(s).}
\label{fig:system-diagram}
\end{figure*}

\section{Introduction}
Industrial manufacturing relies on computer-aided design (CAD) to create products ranging from furniture to Formula 1 cars \cite{ikea2024cad, stoermann2023formulaone}. Software tools such as OnShape \cite{onshape}, SolidWorks \cite{solidworks}, and FreeCAD \cite{freecad} enable engineers to design 3D objects in a GUI by sequentially applying operations (e.g., sketch, extrude, fillet, etc.) that transform 2D sketches and 3D objects into a final desired form. This series of operations, each parameterized by discrete and/or continuous values representing, for example, the length of an extrusion, defines a \textit{CAD model} which can be fabricated.

While CAD tools have become indispensable for manufacturing---increasing productivity, design precision, and product quality---they can still be tedious to use \cite{robertson1991caduse, deng2023cadexperiment}. As a result, the graphics and AI communities have sought to automate the generation of CAD models by training deep generative models to reverse engineer them from images, text descriptions, 3D point clouds, and more \cite{you2025img2cad, khan2024textcad, khan2024cadsignet, chen2025cadcrafter, lambourne2021brepnet, li2025cadllama, rukhovich2025cadrecodereverseengineeringcad, xu2025cadmllmunifyingmultimodalityconditionedcad, xie2025texttocadquerynewparadigmcad}. To overcome a lack of training data, more recent work has built agentic systems where an agent can write code, compile the code to a CAD model, visualize the model in e.g. FreeCAD, and use the visual feedback (as well as compiler errors and other signals) to iterate on its design until the CAD model is sufficiently similar to the input \cite{alrashedy2025generating, mallis2024CADAssistantTV, fan2025caddesignerconceptualdesigncad, anonymous2026seekcad}.

However, in spite of these techniques' state-of-the-art performance on CAD benchmarks \cite{wu2021deepcad, willis2021fusion}, they suffer from a key limitation: the systems are capable of generating static CAD models, \textit{but they cannot model complex assemblies with joints and moving parts}. Put more simply, these systems cannot create a functional pair of scissors. Prior work can create a 3D model for the scissor blades and handles, but it cannot model the revolute joint which allows the blades to move in a slicing motion. Passing ``the scissors test" is crucial if Agent-Aided Design (AAD) is to be taken seriously as a tool for industrial manufacturing.

Generating dynamic 3D assemblies with moving parts is more difficult than generating static CAD models. For starters, the agentic system needs to specify the joints that connect each part in the assembly. Unlike prior work, this includes specifying each joint's degree(s) of freedom to allow for dynamic part interactions. This is made especially challenging, given Vision Language Models' (VLMs) known limitations in spatial reasoning \cite{stogiannidis2025mindgapbenchmarkingspatial}. Furthermore, in order for the agentic system to compile its design, it must have access to a 3D assembly constraint solver. In our initial experiments, we found that existing open-source solvers had limitations due to the fact that they were designed with a human end-user in mind. For example, these solvers used Euler angles instead of quaternions because they assumed that a human could flip the part in a GUI if the solver oriented a part in an anti-parallel fashion. To maximize the solver's utility for an agent, we ultimately had to modify the solver to use quaternions and enhance its error messages to be generally more informative.

In this paper, we present a demo of our work on \system{} as a first step towards building an agentic system for Agent-Aided Design of complex 3D assemblies. In addition to generating CAD models for static parts and assemblies, \system{} is capable of generating dynamic CAD models with free joints and moving parts. At a high-level, the system consists of an agent which is instructed on how to write assemblies via an intermediate representation specified in JSON. This representation is then compiled to a 3D assembly, with compilation errors and renderings generated in FreeCAD being fed back into the agent. Given this feedback, the agent iterates on its design and performs a series of checks to assess whether the design is internally consistent (e.g. that each part and joint looks correct in isolation). Finally, once the agent is satisfied that the assembly meets the input specification, the final 3D model is returned. A system diagram for \system{} is shown in \Cref{fig:system-diagram}.

In order to overcome deficiencies in VLMs' spatial reasoning, we modified FreeCAD to render each part in our assembly with a unique edge color and texture. Additionally, we rendered each instance of the same part with a unique text identifier (e.g. ``Blade1", ``Blade2", etc.). This visual scheme provided the agent with deterministic, stable identifiers which improved its ability to reference specific edges in joint definitions. To improve our agent's efficiency in generating correct outputs, we modified the OndselSolver \cite{ondselsolver}, an open-source Multibody Dynamics Solver, to improve its utility for the agent by, for example, incorporating quaternions and providing more detailed error messages about inconsistent constraints in the assembly and convergence failures. Our approach was inspired by prior work which demonstrated that having \textit{strong verifiers}, i.e. functions which provide robust error / feedback signals to an agentic system \cite{khattab2024dspy, agrawal2025gepareflectivepromptevolution, cheng2025letbarbariansinai, hamadanian2025glia, anonymous2026shinkaevolve, wei2026multiobjectiveagenticrewritesunstructured, russo2025abacuscostbasedoptimizersemantic}, can improve performance (perhaps even more than making changes to the agent itself).

We demonstrate a prototype of \system{} and use it to perform two evaluations. First, we show that \system{} is able to go beyond the capabilities of prior work by generating a functional pair of scissors from two input images. Second, we demonstrate that \system{}'s design also supports constructing static 3D assemblies, some of which were generated by prior work and one of which could not be generated without our modifications to FreeCAD and the 3D constraint solver.



\section{Related Work}
\noindent

\noindent
\textbf{AI for Computer-Aided Design (CAD).} Prior work in the graphics and AI communities has focused on fully (or partially) automating the generation of parametric CAD models. Early work trained generative models to transform input 3D point clouds \cite{khan2024cadsignet}, boundary representations (i.e., B-rep) \cite{lambourne2021brepnet}, images \cite{you2025img2cad, chen2025cadcrafter}, and text descriptions \cite{khan2024textcad} into a representative parametric CAD model. Generally speaking, these techniques required learning from labeled CAD datasets \cite{wu2021deepcad, willis2021fusion} which were hard to come by and had sparse coverage for many types of objects. This led to limitations in generating more complex and / or out-of-distribution objects \cite{khan2024textcad, you2025img2cad}. More recent work has focused on leveraging pretrained large language models (LLMs) to generate CAD models. CAD-LLama \cite{li2025cadllama}, CAD-GPT \cite{wang2025cadgpt}, and others \cite{rukhovich2025cadrecodereverseengineeringcad, xu2025cadmllmunifyingmultimodalityconditionedcad, xie2025texttocadquerynewparadigmcad} finetuned (multimodal) LLMs (MLLMs) to produce code that could be compiled to a 3D model of an input point cloud, image, and / or text string. These works benefited from LLMs' ability to write code matching an input specification, however they still required constructing a training dataset for finetuning.

To move towards a training-free paradigm, the latest systems have taken an agentic approach to generating CAD models. CADCodeVerify \cite{alrashedy2025generating}, CAD-Assistant \cite{mallis2024CADAssistantTV}, CADDesigner \cite{fan2025caddesignerconceptualdesigncad}, and others \cite{anonymous2026seekcad} have placed MLLMs inside of feedback loops in which the MLLM (i.e. agent) (1) writes code to generate a CAD model, (2) gets feedback in the form of compilation errors and object renderings, and (3) reasons about its error(s) to produce a new CAD model or terminates once it believes it has successfully constructed the object. In contrast to prior work, these systems do not require training (beyond the base model's pretraining) and they take an iterative approach to constructing the final CAD model. CAD-Assistant \cite{mallis2024CADAssistantTV} is most similar to \system{}. It generates CAD code actions which are executed in FreeCAD via its Python API, and then uses a set of tools to produce visual feedback from renderings of the object. (CADDesigner also sounds similar, but it is a preprint with no open-sourced code). \system{} differs from CAD-Assistant in its additional focus on designing assemblies with moving parts. \\

\noindent
\textbf{Verification and Optimization for AI Systems.} A new paradigm for agentic systems has sought to improve their performance by providing them with feedback from strong verifiers. Work on AI-driven research for systems (and algorithms) \cite{cheng2025letbarbariansinai, hamadanian2025glia, anonymous2026shinkaevolve} has demonstrated that coding agents can achieve state-of-the-art performance when they are able to iteratively refine the implementation of a system (or algorithm) with feedback in the form of performance metrics on a predefined evaluation. Related work has demonstrated that strong verifiers can help agents solve math and coding problems \cite{cobbe2021trainingverifierssolvemath, jimenez2024swebench} with autograders and unit-tests providing feedback on intermediate results, respectively. When strong verifiers are not available, labeled data and LLM-based evaluation have also been used to optimize agentic systems. DSPy \cite{khattab2024dspy} optimizes declarative language model programs using optimizers \cite{agrawal2025gepareflectivepromptevolution} which receive signals from a user-defined metrics function (which may use an LLM for e.g. evaluating a summary). In a similar vein, semantic query processing engines such as DocETL \cite{shankar2024docetlagenticqueryrewriting, wei2026multiobjectiveagenticrewritesunstructured}, LOTUS \cite{patel2025semanticoperatorsdeclarativemodel}, Palimpzest \cite{liu2025palimpzest, russo2025abacuscostbasedoptimizersemantic}, and others \cite{russo2025deepresearchnewanalytics, liskowski2025cortexaisqlproductionsql} have optimized semantic operator programs with performance measurements (e.g. cost and latency) as well as quality measurements derived from labeled data, LLMs, and proxy methods. While \system{} is similar in its use of a strong verifier for optimization, the domain of 3D CAD model generation differs significantly from these works.
\section{System Overview}

\noindent
\textbf{Agent Execution.} \system{} takes one or more images and optionally a text description of the desired object as input. These inputs are fed into an agent (e.g., Gemini 3 Flash in our demonstration) whose goal is to generate a CAD model (i.e., a 3D assembly) which could be fabricated to create this object. The agent creates an assembly by writing JSON files to define the parts and joints that make up a 3D assembly.

In our example in \Cref{fig:system-diagram}, the agent is given two photos of a pair of scissors as well as a text description for its input. It then constructs three JSON parts files corresponding to the blade, the handle, and the pivot. To compose these parts into an assembly, the agent writes a JSON assembly file which consists of three sections. First, the agent specifies the parts used in the assembly with pointers to their JSON part file definitions. Second, the agent creates the unique set of \textit{part instances} in the ``links" section. For example, a pair of scissors will have two distinct instances of the handle and blade parts, each of which is defined with its own link. Third, these unique part instances are joined together by a set of ``joints".

In \system{}, the agent joins two links together by specifying their names, the type of joint, which two faces of the links are being joined, which degrees of freedom are fixed vs. free, and (if free) what limits exist on their freedom. Once the agent creates a complete JSON assembly file, \system{} compiles the file, renders it in FreeCAD, and sends it to the 3D assembly constraint solver. FreeCAD produces a rendering of the assembly while the constraint solver may produce error messages indicating flaws in the design. (We discuss each of these tools in more detail below). Finally, the agent incorporates these feedback signals into its next generation of the JSON parts and assembly files. This iterative process continues until the assembly passes the constraint solver and the agent deems the rendering of the assembly to adequately represent the input(s). \\

\noindent
\textbf{Building Strong Verifiers.} The agent within \system{} is provided with two primary tools for obtaining a signal on the correctness of its design. First, the agent is provided with a 3D assembly constraint solver (OndselSolver \cite{ondselsolver}) which takes the JSON assembly file and returns whether its parts could physically be joined into an assembly. \system{} uses an augmented version of the OndeslSolver, because some aspects of the original solver hindered the agent's ability to produce correct assemblies.

For example, due to its use of Euler angles, the solver could return an assembly which oriented a part in an anti-parallel fashion. FreeCAD resolved this issue for humans by presenting them with a button to flip any part in the assembly. However, this confused the agent about the correctness of its design. To make the solver more useful for the agent, we augmented it to use quaternions instead of Euler angles. Additionally, we modified the solver to update parts' positions in FreeCAD even when there was a compilation error. By default FreeCAD will not update the design if there is a compilation error (e.g. because two parts would intersect), however visualizing these errors helped the agent understand its mistakes. Finally, we made the solver's error messages more informative and made its internal invocation of Newton's Method more deterministic.

The second tool provided to \system{}'s agent is an extended version of FreeCAD. The agent uses this tool to generate renderings of its 3D assembly. This helps to identify flaws in its design which can be corrected in subsequent iterations. One key challenge with providing visual feedback is that VLMs are imperfect spatial reasoners \cite{stogiannidis2025mindgapbenchmarkingspatial}. In our early experiments, we found that the agent struggled to specify joints correctly because different part instances could look identical in FreeCAD's renderings. To overcome this issue, we modified FreeCAD to compute a unique color and texture (e.g. solid vs. dotted line) for each face and edge of each part instance. Given these unique colors and textures, we could then ask the agent to specify its joints by writing the colors and textures of the part faces to be joined by a given joint. The addition of this enhancement is ultimately what enabled \system{} to produce a correct assembly for the toddler bed in \Cref{sec:demo}.
\begin{figure}
\includegraphics[width=\linewidth]{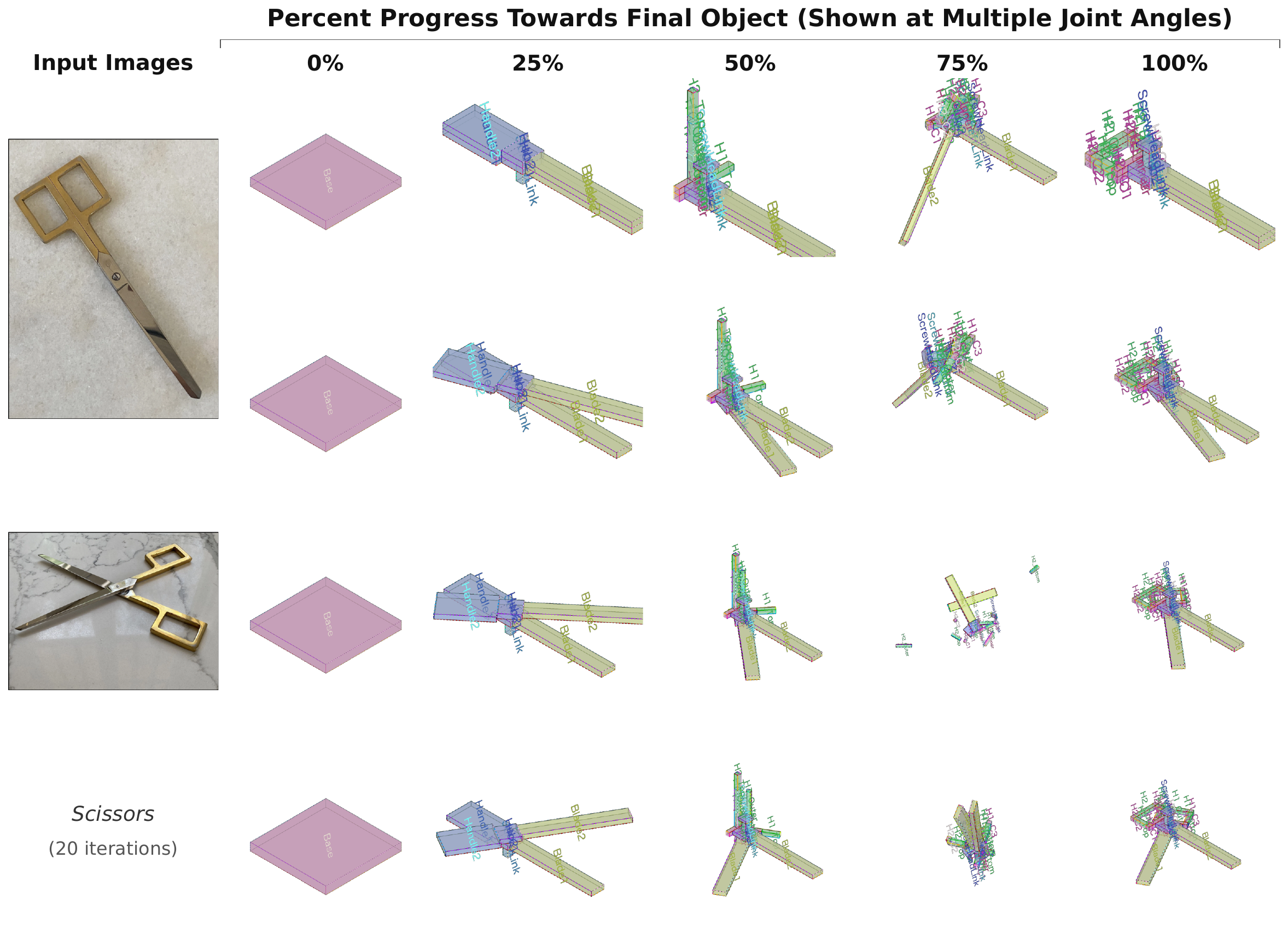}
\caption{Snapshots from our demonstration of \system{} creating a dynamic 3D assembly for a pair of scissors. Each column corresponds to a single iteration of the object as it was generated. Each row shows the object at a different angle of the revolute joint (0, 20, 40, and 60 degrees). \system{} starts out with a base rectangle and by the end of our demonstration it has created a functional pair of scissors.}
\Description{Snapshots from our demonstration of AADvark creating a dynamic 3D assembly for a pair of scissors. Each column corresponds to a single iteration of the object as it was generated. Each row shows the object at a different angle of the revolute joint (0, 20, 40, and 60 degrees). AADvark starts out with a base rectangle and by the end of our demonstration it has created a functional pair of scissors.}
\label{fig:scissors}
\end{figure}

\section{Demo Scenario}
\label{sec:demo}
\textbf{Generating Dynamic CAD Models.} For our demonstration, we tested whether \system{} could generate a dynamic CAD model for a pair of scissors. The input to \system{} was two images. The first image showed a pair of scissors in a closed state and the second image showed them in an open state (left column of \Cref{fig:scissors}). \system{} initialized an assembly with a simple base rectangle and then began to iteratively modify the assembly.

After four iterations (the 20\% progress column in \Cref{fig:scissors}) \system{} already had modified the scissors assembly to resemble the input image(s), including by properly constructing the revolute joint to enable a cutting motion. However, the handles did not have holes for fingers. \system{} then spent the majority of its iterations modifying the handles to contain holes for fingers, before finally accomplishing this task in iteration 20. On average, each iteration took 745 seconds, cost \$0.79, and processed 914k input tokens and 111k output (and thinking) tokens across 23.4 LLM calls. In total, it took 4.14 hours, cost \$15.85, and processed 18.2M input tokens and 2.2M output tokens across 468 LLM calls. \\

\begin{figure}
\includegraphics[width=\linewidth]{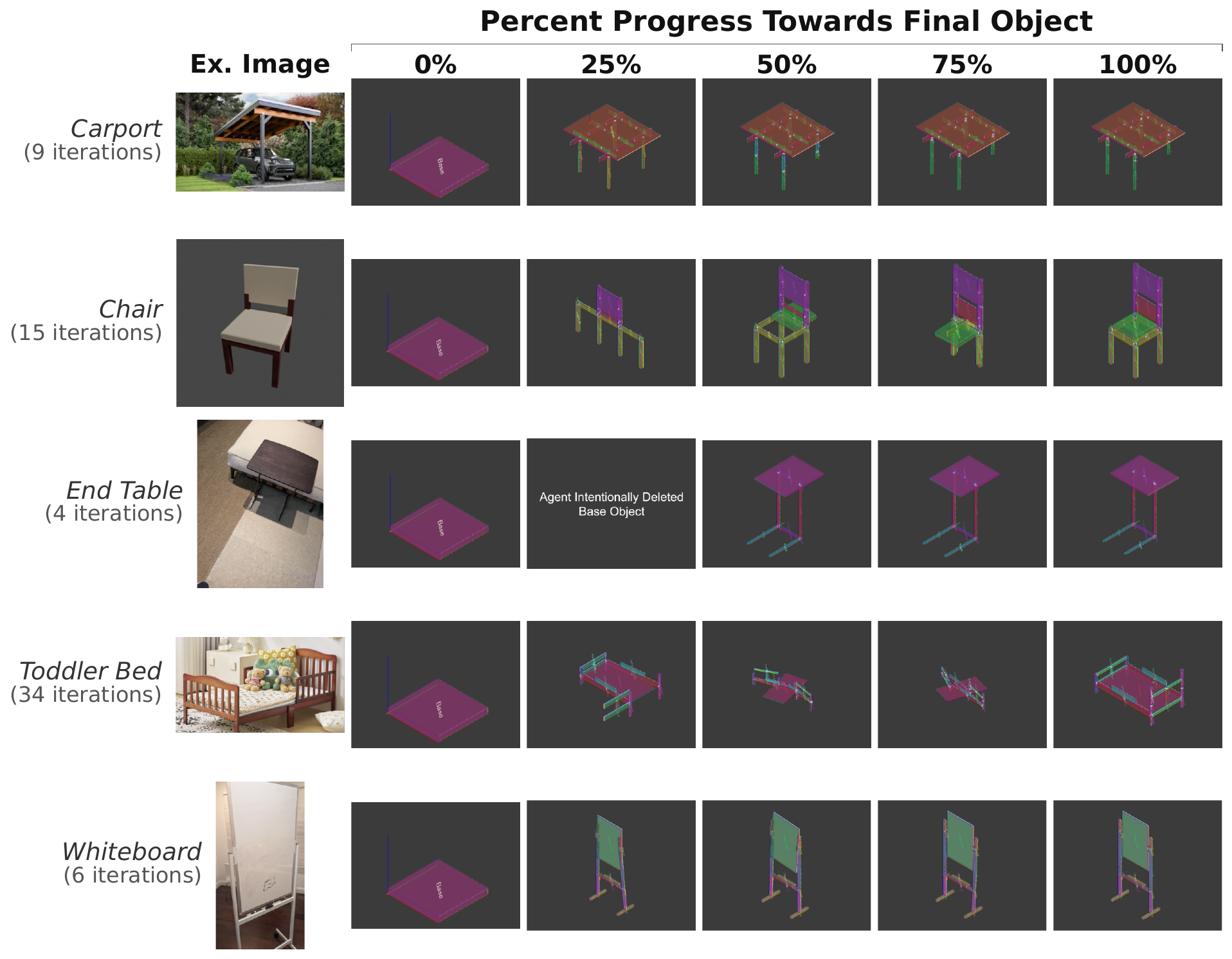}
\caption{Illustration of \system{}'s ability to create 3D assemblies for various objects in FreeCAD. Each object begins as a single base rectangular prism before \system{} iteratively adds parts and joints to construct the final CAD model.}
\Description{Illustration of AADvark's ability to create 3D assemblies for various objects in FreeCAD. Each object begins as a single base rectangular prism before AADvark iteratively adds parts and joints to construct the final CAD model.}
\label{fig:static-models}
\end{figure}

\noindent \textbf{Generating Static CAD Models.} As a second test, we sought to demonstrate that \system{} could also generate 3D assemblies for a diverse set of static objects, some of which were generated by prior work \cite{you2025img2cad}. We provided the system with images of a toddler bed, chair, end table, carport, and a standing dry erase board. For each object we provided 1-3 input images. To probe \system{}'s ability to handle natural language specifications, we additionally provided an LLM-generated design document for how to build a toddler bed. The results from our experiments are shown in \Cref{fig:static-models}

Overall, \system{} was able to generate a CAD model for each input object within 4 - 34 iterations. Once again, \system{} initialized each assembly with a base rectangular prism. For the carport, end table, and whiteboard, \system{} was able to generate a 3D assembly which resembled the input within 1-3 iterations. From that point onward, \system{} simply needed to tweak the size of various parts and the joints between them in order to produce a final assembly which satisfied the constraint solver and looked visually similar to the input (as judged by a VLM). For the toddler bed and chair, \system{} required a larger number of iterations to work through the more complicated set of part interactions (especially for the toddler bed). In our experimentation, we found that \system{} was only able to successfully construct the toddler bed after we augmented the assembly renderings to include unique colors for the different faces and edges of each part instance.

\section{Limitations}
While \system{} represents a promising first step towards Agent-Aided Design of dynamic CAD models, it does have limitations. For now, the system can only use rectangular prisms for parts and the only dynamic joint which it supports is a revolute joint. In the future, we intend to expand the set of parts and joints that \system{} can create to allow it to design any object that FreeCAD can support. Finally, due to the non-deterministic nature of agentic execution, we observed that \system{} sometimes will get stuck on its assembly design. We found that simply restarting the agent can often times resolve the issue, but in the future we aim to investigate the underlying causes of the agent getting stuck.

\begin{acks}
We are grateful for the support from the DARPA ASKEM Award HR00112220042, the ARPA-H Biomedical Data Fabric project, NSF DBI 2327954, a grant from Liberty Mutual, a Google Research Award, and the Amazon Research Award. Additionally, our work has been supported by contributions from Amazon, Google, and Intel as part of the MIT Data Systems and AI Lab (DSAIL) at MIT, along with NSF IIS 1900933. This research was sponsored by the United States Air Force Research Laboratory and the Department of the Air Force Artificial Intelligence Accelerator and was accomplished under Cooperative Agreement Number FA8750-19-2-1000. The views and conclusions contained in this document are those of the authors and should not be interpreted as representing the official policies, either expressed or implied, of the Department of the Air Force or the U.S. Government. The U.S. Government is authorized to reproduce and distribute reprints for Government purposes notwithstanding any copyright notation herein.
\end{acks}

\bibliographystyle{ACM-Reference-Format}
\bibliography{main}


\end{document}